\documentclass{llncs}
\usepackage{graphicx}
\usepackage{adjustbox}
\usepackage{latexsym}
\usepackage[pdftex]{hyperref}

\usepackage[utf8]{inputenc}

\usepackage{comment}

\usepackage[]{todonotes}
\newcommand{\assignment}[1]{}

\begin{document}


\title{Deep Neural Networks for Approximating Stream Reasoning with C-SPARQL}

\author{
Ricardo Ferreira
\and
Carolina Lopes
\and
Ricardo Gon\c calves
\and
Matthias Knorr
\and
Ludwig Krippahl
\and
Jo\~ao Leite
}

\institute{
NOVA LINCS \& Departamento de Inform\'atica, Universidade Nova de Lisboa
}

\maketitle

\begin{abstract}

The amount of information produced, whether by newspapers, blogs and social networks, or by monitoring systems, is increasing rapidly.
Processing all this data in real-time, while taking into consideration advanced knowledge about the problem domain, is challenging, but required in scenarios where assessing potential risks in a timely fashion is critical.
C-SPARQL, a language for continuous queries over streams of RDF data, is one of the more prominent approaches in stream reasoning that provides such continuous inference capabilities over dynamic data that go beyond mere stream processing.
However, it has been shown that, in the presence of huge amounts of data, C-SPARQL may not be able to answer queries in time, in particular when the frequency of incoming data is higher than the time required for reasoning with that data.
In this paper, we investigate whether reasoning with C-SPARQL can be approximated using Recurrent Neural Networks and Convolutional Neural Networks, two neural network architectures that have been shown to be well-suited for time series forecasting and time series classification, to leverage on their higher processing speed once the network has been trained.
We consider a variety of different kinds of queries and obtain overall positive results with high accuracies while improving processing time often by several orders of magnitude.
\end{abstract}

\section{Introduction}
\label{intro}

Large amounts of data are constantly being produced, whether by newspapers, blogs and social networks, or by monitoring systems (such as traffic sensors, financial market prediction, weather forecasting, etc.) \cite{margara2014streaming}. 
For such data streams, it is often necessary to be able to infer new information with high efficiency in real time. 
For example, observing data about a patient's health status, diet and physical activity can help to anticipate health problems and request medical support in case of an emergency.
Also monitoring traffic in a city area can help react to problems, such as traffic jams or accidents, in real time allowing to reroute traffic to  improve travel time and reduce environmental impact.

Data Stream Management Systems (DSMS) and Complex Event Processors (CEP) tackle this problem \cite{cugola2012processing}, where the former allow continuous querying over a data stream and the latter aim at identifying patterns of events that occur in a data stream.
However, these systems cannot handle situations when the data is heterogeneous and it is necessary to integrate background knowledge, such as the patient's record, data on medication, or general medical knowledge expressed in an ontology, and perform more complex reasoning tasks.

Stream reasoning aims at overcoming these limitations \cite{della2009s,dell2017stream,margara2014streaming}  and one of the more prominent approaches is C-SPARQL (Continuous SPARQL) \cite{barbieri2010incremental,barbieri2010c}, a language for continuously querying over streams, that combines the features of DSMS and CEP, and the ability to incorporate background knowledge.
\mbox{C-SPARQL} builds on language standards for the Semantic Web whose development has been driven by the World Wide Web Consortium (W3C), namely, on the Resource Description Framework (RDF) \cite{miller1998introduction}, a standard model of data exchange on the Web which has led to the development of Linked Open Data \cite{HeathB11} with a large amount of structured and interconnected data,\footnote{https://lod-cloud.net/} and SPARQL the query language for querying over RDF data \cite{sparql}.
In more detail, C-SPARQL is able to process queries over various RDF streams simultaneously, employing so-called windows that focus only on a limited (recent) portion of the stream, taking into account background knowledge in the form of RDF graphs. 
Query answers can be variable bindings or again RDF graphs, and both of these even in the form of streams, thus C-SPARQL is capable of updating knowledge bases as new information arrives.

While C-SPARQL is thus in principle well-suited to perform stream reasoning, these advanced reasoning capabilities come at a price \cite{ren2016measuring}.
It has been shown that there is a limit on the amount of triples the system is capable of processing per second, which varies depending on the complexity of the considered query, and which may be prohibitive in real world scenarios.
In fact, at higher rates where the amount of triples per second is superior than the limit, C-SPARQL provides erroneous answers.

It has been argued that in the face of huge amounts of data, sound and complete reasoning can be considered as a gold standard, but for obtaining answers in a timely fashion, approximate methods need to be applied \cite{HitzlerH10}.
These approximate methods include approaches based on machine learning, such as Neural Networks (NNs) \cite{haykin1994neural}, which are able to learn and generalize patterns from a given data set, making them, once trained, applicable to unseen situations with a considerably higher processing speed and robust to noisy data \cite{HitzlerBES20}.
These methods have gained further interest with the advent of Deep Neural Networks \cite{GoodfellowBC16}, which are behind a variety of substantial recent advances in Artificial Intelligence, for example, in speech and visual recognition oe vehicle control, and which allow to detect considerably more sophisticated patterns in a data set.

A few solutions to such deductive reasoning using Deep Learning have appeared within the field of neural-symbolic integration \cite{BesoldEtAl17,HitzlerBES20}.
Namely, Makni and Hendler \cite{MakniH19} propose noise-tolerant reasoning for RDF(S) knowledge graphs \cite{rdfs}, Hohenecker and Lukasiewicz \cite{HoheneckerL20} introduce Recursive Reasoning Networks for OWL 2 RL reasoning \cite{owlProfiles}, and Ebrahimi et al.\ \cite{EbrahimiEtAl18} tackle learning of simple deductive RDFS reasoning.
However, none of these approaches takes streaming data into account, making them unsuitable in the scenarios where reacting to temporal sequences of events is required. 

In this paper, we investigate whether reasoning with C-SPARQL can be approximated using Deep Learning.
We consider Recurrent Neural Networks (RNNs) \cite{qin2017dual} and Convolutional Neural Networks (CNN) \cite{TSC}, two neural network architectures that have been shown to be well-suited for time series forecasting and time series classification \cite{BianchiSLJ21,fawaz2019deep}.
Using a data set containing sensor data on traffic, pollution and weather conditions, we consider different kinds of queries aiming to cover different features within the expressiveness C-SPARQL offers, and we generate the target labels for the training set using C-SPARQL itself to avoid the cost of manually labeling the data.
We are able to show that such approximate reasoning is indeed possible and obtain overall positive results with high accuracies while improving processing time often by several orders of magnitude.
We also provide considerations on which of the two arquitectures is more suitable in which situation.


\section{Background}
\label{state-of-art}

In this section, we recall relevant notions to facilitate the reading of the remaining material, namely on C-SPARQL, and the two kinds of neural networks we consider, recurrent and convolutional neural networks.

\subsection{C-SPARQL}

C-SPARQL (Continuous SPARQL) \cite{barbieri2010incremental,barbieri2010c} is a declarative query language that combines the features of DSMS and CEP to continuously query over streams of data taking into account background knowledge.
C-SPARQL builds on the Resource Description Framework (RDF) \cite{miller1998introduction}, a standard model of data exchange on the Web and SPARQL the query language for querying over RDF data \cite{sparql}.

To represent continuous streams of data, RDF streams are introduced \cite{barbieri2010incremental}.
An \emph{RDF stream} is an ordered sequence of pairs, where each pair consists of an RDF triple $ \langle subject,predicate,object\rangle$ and its timestamp $ \tau$. 
Subsequent timestamps $\tau_i$ and $\tau_{i+1}$  are monotonically non-decreasing ($\tau_i\leq\tau_{i+1}$), i.e., any (unbounded, though finite) number of consecutive triples can have the same timestamp, but they still occur sequentially according to the given order. 

To be able to process such streaming data, C-SPARQL applies \emph{windows} to delimit a finite amount of triples to be considered, which aligns with the idea that we cannot consider the entire stream when reasoning (nor store it).
Such windows can be \emph{physical}, i.e., a specific number of triples is selected, or \emph{logical}, i.e., the triples that occur in a certain interval of time are selected.
The latter can be \emph{sliding} windows, when they are progressively advancing with a time step smaller than their interval, or \emph{non-overlapping} (also called \textit{tumbling}) when they are advancing with exactly their time interval at each iteration \cite{barbieri2010c}.

Continuous queries in C-SPARQL extend SPARQL queries \cite{sparql} with the necessary features to handle streams of data.
More concisely, a C-SPARQL query starts with a registration statement to be able to produce continuous output in the form of variable bindings, tables or graphs, also indicating at which frequency the query is processed.
A basic C-SPARQL query then contains \textit{SELECT}, \textit{FROM} and \textit{WHERE} statements, where the \textit{SELECT} statement indicates the variables one is interested in, the \textit{FROM} statement the IRI of the stream considered, including the definition of the kind of applied window, and the \textit{WHERE} statement a condition for the query.
In addition, C-SPARQL queries also permit the usage of the following advanced characteristics:
\begin{itemize}
\item Aggregate functions, namely, count, sum, average, minimum and maximum, which additionally allow grouping results as well as subsequent filters on the aggregated data;	
\item Incorporation of static background knowledge in the form of external RDF documents that can be referred to in the query (optionally introduced in the prefix statement);
\item Usage of the timestamp function to be able to compare the time of occurrence of different events in the window of the stream;
\item Querying various streams simultaneously;
\item Stream the results of the continuous query, in addition to returning the variable bindings and graphs.
\end{itemize}
For the concise description of the syntax of SPARQL queries, and their formal semantics based on mappings we refer to \cite{barbieri2010c}.

It should be noted that, as shown in \cite{ren2016measuring}, the query execution time depends on the complexity of the query and increases linearly with the growth of the window size and the size of the static background knowledge, effectively limiting the number of triples that can be processed per unit of time, resulting in wrong answers if this threshold is passed.

\subsection{Neural Networks for Time Series Classification}

We assume a basic understanding of (deep) neural networks and how they can be trained using supervised learning \cite{GoodfellowBC16,haykin1994neural}.
Here, we provide an overview on the two network architechtures particularly well-suited for time series classification.

\subsubsection{Recurrent Neural Networks}

Recurrent neural networks (RNNs) are a type of deep neural networks specially designed for sequence modelling, that have received a great amount of attention due to their flexibility in capturing nonlinear relationships \cite{qin2017dual}. 
An RNN is very similar to a feedforward neural network with the exception that, in RNNs, the output of the recurrent layer is passed as input to that same layer in the next time step. 
At each time step, the neurons receive an input vector and the output vector from the previous time step. RNNs have had great success in forecasting and classifying time series.
However, they suffer from the vanishing gradient problem \cite{PascanuMB12}, which causes the RNN to lose the ability to "learn more" at a certain point and results in difficulties capturing long-term dependencies. 
To overcome both problems, more sophisticated architectures were introduced such as Long Short-Term Memory Units (LSTM) \cite{hochreiter1997long}, and Gated Recurrent Units (GRU) \cite{cho2014learning}, which are a simplified version of LSTM, aimed at being useful for smaller datasets.

\subsubsection{Convolutional Neural Networks}

Convolutional Neural Networks (CNNs) have first been applied very successfully in the context of image recognition, and have shown impressive results when dealing with
Time Series Classification \cite{fawaz2019deep}.
Their distinguishing characteristic is the usage of convolutional and pooling layers for detecting patterns \cite{TSC}.

A convolutional layer aims at recognizing patterns that exist in the data.
This layer applies a finite number of filters to the input, creating as output, for each filter, a feature map with the patterns detected by that filter. The values of the kernel matrix used in the convolution operation for each filter are adjusted during training, so that the network can learn the patterns it needs to find.
Pooling layers then allow us to reduce the dimensions of such data by joining the outputs from a portion of the neurons in the previous layer into a single neuron of the following layer, using, e.g., \emph{Max pooling} where the highest value is selected.
This way, the data is compressed, simplifying the result without losing the most relevant information.
Combining these layers leads to a gradual reduction in the amount of information to be processed along the network and requires fewer parameters than an equivalent fully connected network.

\section{Methodology}
\label{methodology}

In order to test our hypothesis that stream reasoning with C-SPARQL can be approximated using RNNs and CNNs, we designed and executed a series of experiments, whose rationale is explained in this section.

In general, each experiment consists of performing the following two steps.

\begin{enumerate}
	\item Formulate a C-SPARQL query using different combinations of features (mentioned in Section~\ref{state-of-art}) and execute this query using C-SPARQL to obtain the correct query answers;
	\item Use an encoding of the original data together with the stream reasoner's answers for developing, training, and testing various models for RNNs and CNNs to determine, for each of them, the one with the best performance in terms of approximation of reasoning.
\end{enumerate}

Both steps are detailed next after introducing the test data we have used.

\subsection{Dataset} \label{D}

To perform our experiments, we chose a publicly available dataset \footnote{\url{http://iot.ee.surrey.ac.uk:8080/datasets.html}} that gathers data on traffic, pollution and weather conditions by 449 sensors distributed throughout the city of Aarhus in Denmark.
The data collected by the sensors is diverse, including average speed, number of vehicles detected, measurements of, for example, carbon, sulfur, ozone, nitrogen, as well as data on temperature, pressure and humidity.

To aid our experiments, we created an event processor that returns discretized events from the data in the dataset.
The benefits of this are two-fold.
Given the known limitations in terms of capacity of processing large amounts of data of \mbox{C-SPARQL}, we can perform simple event detection a-priori to reduce the amount of data to be processed with C-SPARQL and at the same time avoid that it be applied for simple event detection where it would not be necessary in the first place.
Such pre-processing also facilitates encoding the streaming data as input for the neural networks, and, thus, in addition allows for a fairer comparison of processing time, as both use the same pre-processed set of input data.

This event processor is a Python script\footnote{\url{https://github.com/CarolinaMagLopes/Deep-Neural-Networks-for-C-SPARQL}.} that creates certain events based on the present data, namely \textit{movement, no$_-$movement, normal air, low carbon, high carbon, low nitrogen, high nitrogen, low sulfure} and \textit{high sulfure}.
For example, if a sensor does not detect any vehicles at a given time instant, the event detector would return the \textit{no$_-$movement} event, and if the carbon values are above normal (based on average values), the {\it high$_-$carbon} event.
In addition, to allow for more interesting queries involving advanced reasoning, we divided the city (and thus the sensors) into 10 sectors with the aim to allow comparisons between different areas.
To balance the amount of data per sector, we only considered 15 sensors per sector, and condensed the time step in the original dataset from 5 minutes to 1 minute.
Thus, all computed events are associated with their sector and a timestamp, given in minutes.
The dataset resulting from this process contains 17532 samples of sliding windows (with a time step of 1 minute) of size 5 corresponding to 5 minutes.

\subsection{C-SPARQL Queries}

The queries we designed aim to leverage the expressive features C-SPARQL offers.
In our experiments, we thus created queries of varying complexity using different combinations of the features, most notably aggregations, background knowledge and time comparisons, incorporating combinatorics over the different sectors, which is not easily coverable by a stream processor.

Considering that Neural Networks are not able to straightforwardly provide RDF triples as answers, we only took into account queries that return answers that can be encoded into a constant number of neurons, i.e., queries that return a Boolean answer (ASK operator) or a fixed number of answers.
For similar reasons, we also abstained from considering C-SPARQL queries that return a stream of RDF data.
This is not a major limitation and the queries we present in Section~\ref{results} are expressive and showcase a wide variety of possible use cases.

To process C-SPARQL queries we have used the Ready-To-Go Pack.\footnote{ \url{http://streamreasoning.org/resources/c-sparql}} 
When processing a query, time and memory usage were measured.
To avoid that \mbox{C-SPARQL} starts giving wrong answers when passing the threshold of processable amount of data per time step, we added sufficient delay to the query processing (as determined in prior tests and higher than any processing times), without affecting the measured processing times themselves.

\subsection{Training RNNs and CNNs}

The result of running the queries are windows of data together with the corresponding query result.
To be able to train a neural network with this data, we needed to encode the data accordingly.
To cover the 9 events in 10 sectors, a matrix of size (\textit{9}$\times$\textit{10}) has been used.
For the case of RNNs, a window of 5 minutes thus results in an input of 5 such matrices, whereas for CNNs an additional dimension of 5 is added, hence each window corresponds to a matrix of size (\textit{9}$\times$\textit{10}$\times$\textit{5}).

Then for each query, varying architectures with different numbers (and kinds) of layers and neurons were designed, trained, and tested for both CNNs as well as LSTM and GRU architectures, the RNNs, we considered here, using also Dropout and Gaussian Noise to avoid overfitting of the networks.
The quality of their reasoning approximation has been measured using the accuracy (the fraction of correct answers) obtained with the test set and from training.

From the 17532 samples, 1532 were reserved for testing and 16000 were used for training and validation, typically with 10\% of these being used in the validation set.
The number of epochs used for training varies depending on the used network and the query, corresponding in all cases to the best results achieved while avoiding overfitting.
For RNNs, this varies between 100 and 1000 epochs, for CNN, 50 epochs provided the best results.

The networks were developed and tested using Python3, namely using \textit{TensorFlow}\footnote{ \url{https://www.tensorflow.org/}}, \textit{Keras}\footnote{ \url{https://keras.io/}} and \textit{SciKit-Learn} \footnote{\url{https://scikit-learn.org/stable/}}, where
Keras in particular provides implementations of CNNs, LSTM and GRU architectures.

\section{Experiments and Results}
\label{results}

In this section, we present the different experiments carried out with the aim of assessing to what extent neural networks are able to approximate reasoning with C-SPARQL. 
Among the many different queries that we tested, we have chosen several representatives of certain combinations of the features of C-SPARQL (cf.~Section~\ref{state-of-art}), and grouped them together in subsections according to the chosen features to facilitate the reading.

In each case, we indicate the representative queries and the processing times of C-SPARQL, RNNs and CNNs, as well as the resulting accuracy for the test set for both network architectures.
Please note that w.r.t.\ RNNs, we often refer directly to LSTM in this section, as they turned out to provide better results than GRU for all the tests of RNNs.

The experiments were run on a computer with an Intel Core i3-3240T processor with 2.90GHz and 4GB of RAM.

As this is common to all experiments, we remark here that running the event processor to create the data set of events only required on average a few milliseconds per sample, meaning that this is irrelevant for the overall processing time of windows covering 5 minutes, i.e., it is insignificant for ensuring whether processing queries is possible in real-time.
We also note that our experiments when processing with C-SPARQL have shown overall that, even for more complicated queries, memory consumption does nor surpass a few hundred MB, which does not constitute a bottle-neck for processing or running the tests, which is why we do not report it individually.
Finally, we also report here that training the networks took on average between 45 minutes to 4.5 hours for RNNs and 4 to 10 minutes for CNNs, depending on the query and the number of epochs to achieve the best results.
While this adds to the time necessary to use a neural network instead of C-SPARQL, it only needs to be done once before applying the network for approximating reasoning. 
Hence, once the network is trained, this does not affect the usage in real time either.

\subsection{Queries with temporal events}\label{subsec:tempEvents}

For the first set of queries, we tested the identification of sequences of temporal events for RDF triples with the timestamp function, which is one of the fundamental characteristics of C-SPARQL. 

Here, we wanted to determine within a window the occurrence, in a sector, of the complex event composed of\par
$t_1$: {\it normal$_-$air} \& {\it no$_-$movement} \qquad 
$t_2$: {\it high$_-$carbon} \& {\it movement}\par 
$t_3$: {\it high$_-$sulfure}\par
\noindent
where $t_1$, $t_2$ and $t_3$ are timestamps such that $t_1<t_2<t_3$.

For this event, we created different kinds of queries to cover the expressiveness of C-SPARQL.
Query 1 selects the sectors where the complex event occurred (using the SELECT operator), while Query 2, tests whether the event occurred in all ten sectors (using the ASK operator).
Query 3 is considerably more complex than the previous two since aggregation is added in the form of COUNT and MAX, aiming to determine the sector with the highest number of occurrences of this complex event.
The obtained results are reported in the following table.\footnote{C-SPARQL processing times vary depending on the number of tuples per window.}
	
\begin{center}
\begin{adjustbox}{max width=0.8\textwidth}
\begin{tabular}{ c | c | c | c | c | c | c | c }
& \multicolumn{1}{c|}{C-SPARQL} & \multicolumn{3}{c|}{LSTM} & \multicolumn{3}{c}{CNN}\\
Results & Time & Train Acc & Test Acc & Time & Train Acc & Test Acc & Time \\
\hline
Query 1& 10-25min & 0.9801 & 0.9780 & 180$\mu s$ & 0.9832 & 0.9976 & 280$\mu s$ \\
Query 2& 10-15min & 0.8760 & 0.8695 & 210$\mu s$ & 0.9898 & 0.9852  & 450$\mu s$ \\
Query 3& 15-30+min & 0.9350 & 0.9311 & 700$\mu s$ & 0.9324 & 0.9222  & 240$\mu s$ \\
\end{tabular}
\end{adjustbox}
\end{center}

We can see that CNNs provide excellent results for both Queries 1 and 2, whereas LSTMs provide excellent results as well for Query 1, but only good results for Query 2 (ASK), which is already observed during training. 
Such difficulties can be circumvented though by counting the results of Query 1, for which the results are excellent and obtain the correct answer with high precision.
For Query 3, which is considerably more complex with the aggregations, very good results are achieved for both kinds of networks.

Overall, we conclude that detecting temporal sequences in stream reasoning using neural networks is feasible with, in general, very good results even for more complex queries.
At the same time, processing with any of the trained networks is at least 6 orders of magnitudes faster than reasoning with C-SPARQL. 
In particular, for the queries considered here, C-SPARQL could not be used with the considered data in real time as the processing time is far higher than the admitted processing time of 1 minute (each window captures 5 minutes, but the time step of the sliding window is 1 minute), thus resulting in incorrect answers in such a setting, unlike the networks that easily permit processing in real time.

\subsection{Queries with Background Knowledge}\label{subsec:onto}

In the second part, we introduced an additional layer of complexity by adding background knowledge in the form of an ontology containing information about the type of sectors ({\it school$_-$area}, {\it urban$_-$area}, etc.), events, called infractions, not allowed in the various types of sectors (for example, the {\it high$_-$carbon} event should not happen in {\it school$_-$area}) and information about adjacency between sectors.
Here, we did not consider temporal comparisons between events, but made use of various aggregation functions.

Query 4 consists in selecting the sectors where more infractions occurred than the average of infractions in the adjacent sectors, thus making use of the topological knowledge, whereas Query 5 asks if the number of sectors, where more infractions occurred than the average of infractions in the adjacent sectors, is greater than or equal to 4, requiring aggregations within an aggregation.

For Query 6, we further increased the complexity of the query and made use of the Property Paths Operators that C-SPARQL provides, namely, among them the '/' operator, where $pred_1/pred_2$ corresponds to the sequence path of $pred_1$ followed by $pred_2$.
This allows to query for sectors where more infractions occurred than the average of infractions in the sectors adjacent to adjacent sectors.
To allow for more meaningful answers, for this query, we increased the number of sectors to 15 choosing sensors in the same way as described in Section~\ref{methodology}, since with the initial 10 sectors, any sector would be close to almost all the others. 

\begin{center}
	\begin{adjustbox}{max width=0.8\textwidth}
		\begin{tabular}{ c | c | c | c | c | c | c | c }
			& \multicolumn{1}{c|}{C-SPARQL} & \multicolumn{3}{c|}{LSTM} & \multicolumn{3}{c}{CNN}\\
			Results & Time & Train Acc & Test Acc & Time & Train Acc & Test Acc & Time \\
			\hline
			Query 4 & 20-40ms & 0.9810 & 0.9775 & 1ms & 0.9797 & 0.9799 & 545$\mu s$ \\
			Query 5 & 20ms & 0.9130 & 0.8930 & 1ms & 0.9180 & 0.9066 & 460$\mu s$ \\
			Query 6 & 30ms & 0.9840 & 0.9771 & 1ms & 0.9795 & 0.9728 & 700$\mu s$ \\
		\end{tabular}
	\end{adjustbox}
\end{center}

Both LSTMs and CNNs perform similarly, showing excellent results for both queries 4 and 6, and only good results for Query 5 (using ASK).
Similar to the solution for Query 2 for LSTMs, we can circumvent this and take advantage of the excellent performance for Query 4 and simply count the results there.
We observe that C-SPARQL reasoning involving ontologies can be captured with LSTMs and CNNs as well.
Notably, for this kind of queries C-SPARQL provides very good processing times that allow its usage in real time, but using the trained networks is still at least 30 and 60 times faster for LSTMs and CNNs, resp.

\subsection{Combining Temporal Events and Background Knowledge}

In the final set of queries, we combine comparisons of temporal events with background knowledge and aggregations to obtain highly sophisticated queries for testing our hypothesis.
We reuse the complex event from Section~\ref{subsec:tempEvents} and the ontology from Section~\ref{subsec:onto} for the two final queries.
Namely, with Query 7 we determine those sectors that have more occurrences of the complex event than the average in the adjacent sectors, and with Query 8, similar to Query 6, those sectors that have more occurrences of the complex event than in the close, but not immediately adjacent sectors.

\begin{center}
	\begin{adjustbox}{max width=0.8\textwidth}
		\begin{tabular}{ c | c | c | c | c | c | c | c }
			& \multicolumn{1}{c|}{C-SPARQL} & \multicolumn{3}{c|}{LSTM} & \multicolumn{3}{c}{CNN}\\
			Results & Time & Train Acc & Test Acc & Time & Train Acc & Test Acc & Time \\
			\hline
			Query 7 & 20-30+min & 0.8990 & 0.8864 & 1ms & 0.8310 & 0.8317 & 460$\mu s$ \\
			Query 8 & 25-30+min & 0.8860 & 0.8718 & 1ms & 0.8584 & 0.8555 & 555$\mu s$ \\
		\end{tabular}
	\end{adjustbox}
\end{center}

We can observe that both kinds of networks achieve good results.
Given the observed results also from training, where our tests with more advanced models would result in overfitting, we believe that this may be due to the fact that for such complicated queries more training data would be necessary, and leave this for future work.
We also note that, here, LSTMs show a slightly better performance. 
We conjecture that this could be related to the fact that for a CNN the entire input is presented as a matrix possibly somewhat obfuscating the temporal aspect, whereas for LSTMs the temporal component of the input is separated, and thus possibly easier to distinguish.
In any case, similar to Section~\ref{subsec:tempEvents}, C-SPARQL cannot be applied in real time, as the processing time by far exceeds the limit, whereas the networks are again six orders of magnitude faster.
Thus, even though there is still space for improvement here in terms of resulting precision for the approximation, given the unsuitability of using \mbox{C-SPARQL} in real time for such queries, our approach is also very promising for such highly complex queries.

\section{Conclusions}
\label{conclusions}

We have investigated whether expressive stream reasoning with C-SPARQL can be approximated using Recurrent and Convolutional Neural Networks, since for more sophisticated queries with higher quantities of data to be processed, \mbox{C-SPARQL} does not process the data fast enough to provide timely answers.

Our experiments on a real data set containing among others data on 
traffic and air pollution show that both RNNs and CNNs are well-suited for this task, as their processing time is vastly superior compared to C-SPARQL, in particular when relative comparisons of temporal events are required, and in many cases, they show excellent or very good results of approximation of reasoning, even when utilizing more complex combinations of C-SPARQL's features.

When comparing both architectures, we noted slightly better results for RNNs ,i.e., LSTMs, when approximating highly complex queries in particular including many temporal comparisons (Query 7 and 8).
On the other hand, CNNs performed a bit better for ASK queries (notably Query 2), which requires counting globally, and training them is notably faster.
Our preliminary conclusion is that CNNs could be more suitable where space perception is more important, while RNNs seem preferable for queries more oriented towards temporal aspects.

Still, C-SPARQL remains more suitable if the frequency of incoming data is below its processing capabilities and when obtaining the correct result is mandatory, or where ad-hoc variants of a query are needed, which would  require re-training the network, as well as in situations when returning answer substitutions where the domain is not previously limited and known, or when returning constructed graphs (as streams).

In terms of future work, it would be interesting to further explore the situations where the networks were not providing excellent results.
This could be tackled by trying to use larger datasets and further test the combinations of features of C-SPARQL that proved more difficult when approximating reasoning. 
Another option would be to consider, e.g., Echo-State Networks (ESN), that have shown promising results in term series classification as well \cite{BianchiSLJ21}.

\paragraph{Acknowledgments}
We thank the anonymous reviewers for their helpful comments and acknowledge support by FCT project RIVER ({PTDC}/{CCI-COM}/{30952}/{2017}) and by FCT project NOVA LINCS ({UIDB}/{04516}/{2020}). 

\bibliographystyle{splncs04}
\bibliography{biblio}

\newpage
\section{Appendix A}
\label{appendixA}


\subsection{SPARQL Queries}

\begin{figure}
\begin{center}
\includegraphics[width=0.9\textwidth]{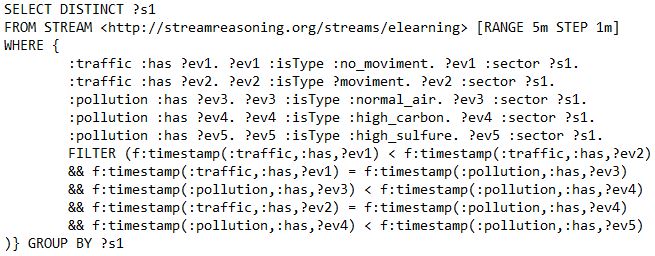}
\caption{C-SPARQL definition for Query 1}
\end{center}
\end{figure}

\begin{figure}
\begin{center}
\includegraphics[width=0.9\textwidth]{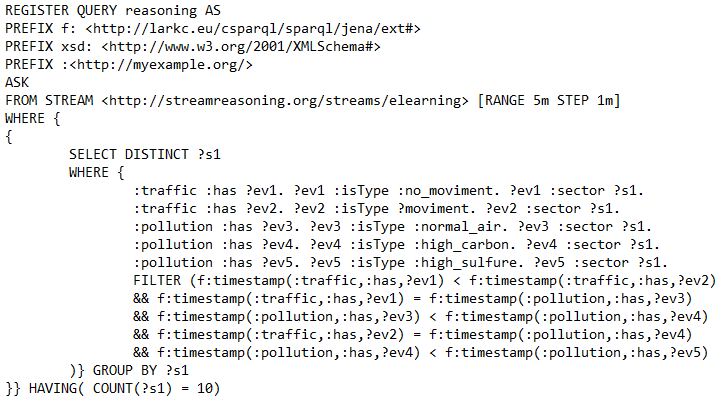}
\caption{C-SPARQL definition for Query 2}
\end{center}
\end{figure}

\begin{figure}
\begin{center}
\includegraphics[width=\textwidth]{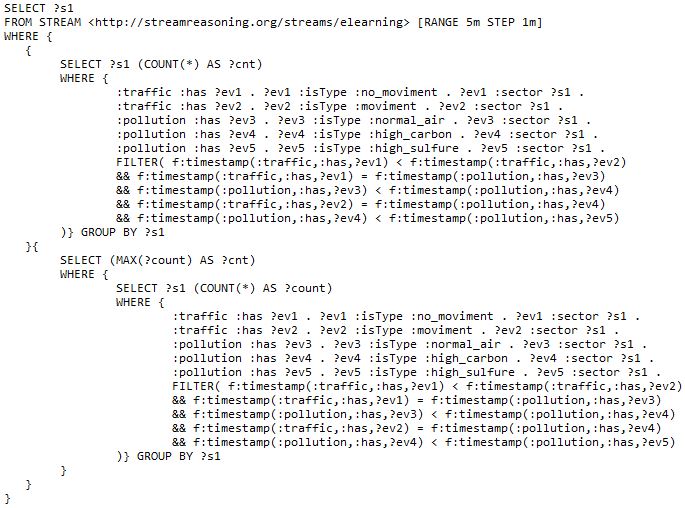}
\caption{C-SPARQL definition for Query 3}
\end{center}
\end{figure}

\begin{figure}
\begin{center}
\includegraphics[width=\textwidth]{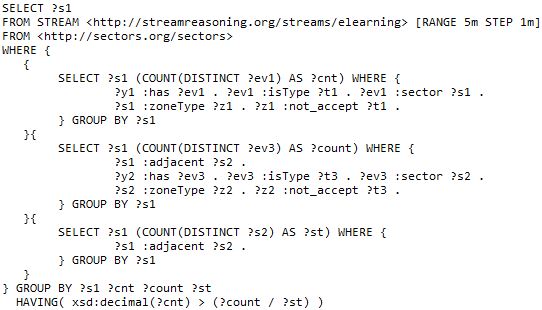}
\caption{C-SPARQL definition for Query 4}
\end{center}
\end{figure}

\begin{figure}
\begin{center}
\includegraphics[width=\textwidth]{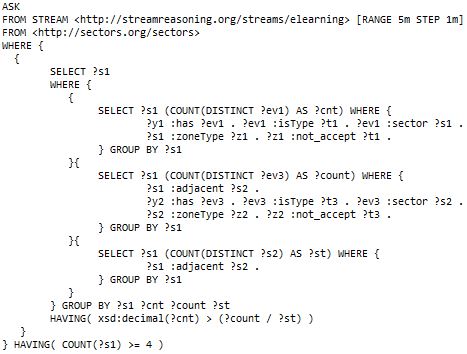}
\caption{C-SPARQL definition for Query 5}
\end{center}
\end{figure}

\begin{figure}
\begin{center}
\includegraphics[width=\textwidth]{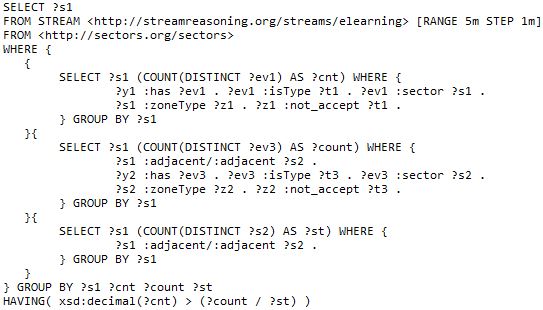}
\caption{C-SPARQL definition for Query 6}
\end{center}
\end{figure}

\begin{figure}
\begin{center}
\includegraphics[width=\textwidth]{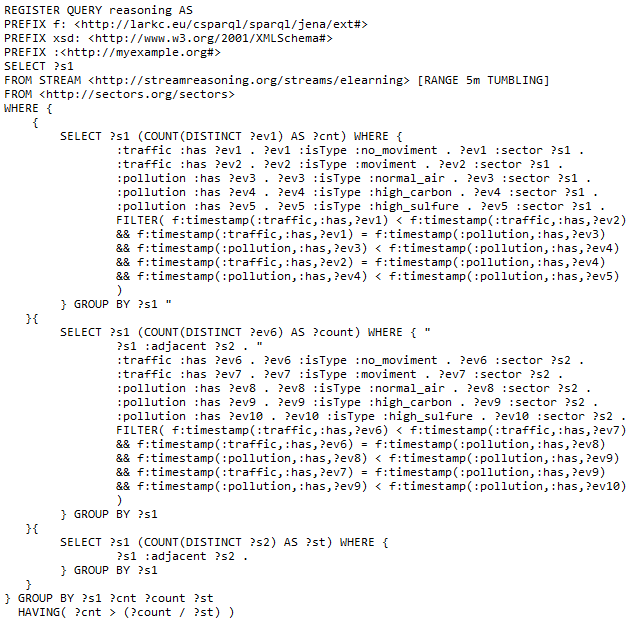}
\caption{C-SPARQL definition for Query 7}
\end{center}
\end{figure}

\begin{figure}
\begin{center}
\includegraphics[width=\textwidth]{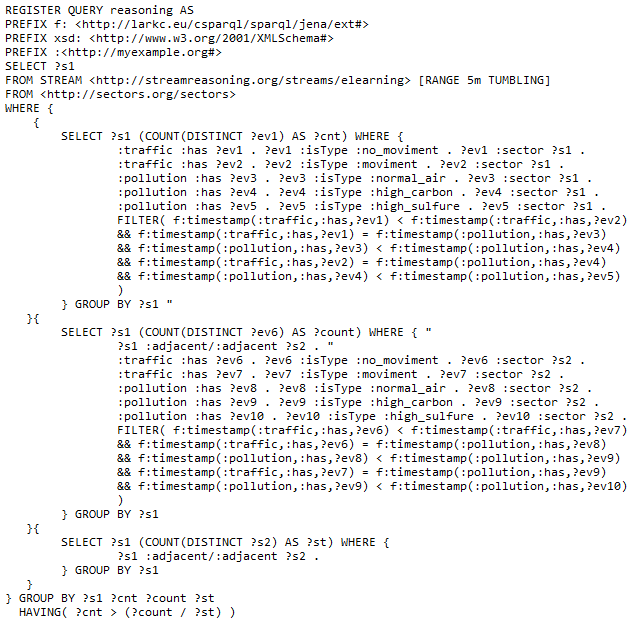}
\caption{C-SPARQL definition for Query 8}
\end{center}
\end{figure}

\newpage
\subsection{Final Network Configurations RNNs}

\noindent
Query 1:
GaussianNoise layer with standard deviation of 0.2, followed by an LSTM layer with 15 neurons, followed by 2 dense layers with 50 and 30 neurons, respectively, and tanh activation, and a final dense layer with 10 neurons with sigmoid activation function. Optimizer: RMSProp.

\noindent
Query 2:
GaussianNoise layer with standard deviation of 0.1, followed by an LSTM layer with 30 neurons, followed by 3 dense layers with 30, 20, and 10 neurons, respectively, and tanh activation, and a final dense layer with 1 neuron with sigmoid activation function. Optimizer: Adam.

\noindent
Query 3:
GaussianNoise layer with standard deviation of 0.3, followed by an LSTM layer with 15 neurons, followed by a Dropout layer of value 0.3, followed by 1 dense layer with 40 neurons and tanh activation and a Dropout Layer of value 0.3, followed by 1 dense layer with 25 neurons and tanh activation and a Dropout Layer of value 0.3, and a final dense layer with 1 neuron with sigmoid activation function. Optimizer: RMSProp.

\noindent
Query 4:
GaussianNoise layer with standard deviation of 0.2, followed by an LSTM layer with 20 neurons, followed by a Dropout layer of value 0.4, followed by 2 dense layers with 30 and 10 neurons, respectively, and tanh activation, and a final dense layer with 10 neurons with sigmoid activation function. Optimizer: Adam.

\noindent
Query 5:
GaussianNoise layer with standard deviation of 0.2, followed by an LSTM layer with 20 neurons, followed by a Dropout layer of value 0.4, followed by 4 dense layers with 50, 30, 20, and 10 neurons, respectively, and tanh activation, and a final dense layer with 1 neuron with sigmoid activation function. Optimizer: Adam.

\noindent
Query 6:
GaussianNoise layer with standard deviation of 0.2, followed by an LSTM layer with 20 neurons, followed by a Dropout layer of value 0.2, followed by 2 dense layers with  30 and 15 neurons, respectively, and tanh activation, and a final dense layer with 15 neurons with sigmoid activation function. Optimizer: RMSProp.

\noindent
Query 7:
GaussianNoise layer with standard deviation of 0.2, followed by an LSTM layer with 50 neurons, followed by a Dropout layer of value 0.3, followed by 1 LSTM layer with 20 neurons, followed by a Dropout Layer of value 0.3, followed by 2 dense layers with 120 and 60 neurons, respectively, and tanh activation, and a final dense layer with 10 neurons with sigmoid activation function. Optimizer: RMSProp.

\noindent
Query 8:
GaussianNoise layer with standard deviation of 0.2, followed by an LSTM layer with 50 neurons, followed by a Dropout layer of value 0.3, followed by an LSTM layer with 20 neurons, followed by a Dropout Layer of value 0.3, followed by 3 dense layers with 120, 80 and 50 neurons, respectively, and tanh activation, and a final dense layer with 15 neurons with sigmoid activation function. Optimizer: RMSProp.

\subsection{Final Network Configurations CNNs}

Query1:
1 Convolutional layer with 32 filters and kernel size of 13, followed by 1 MaxPooling layer with a pool size of 2, followed by
1 Convolutional layer with 16 filters and kernel size of 3, followed by 1 MaxPooling layer with a pool size of 2, followed by
1 Flatten layers followed by 1 Dense layer with 64 neurons with 0.7 of Dropout followed by 1 last dense layer with 10 neurons with a sigmoid function. Optimizer: Adam. Loss Function: Binary Crossentropy.

\noindent
Query2:
1 Convolutional layer with 64 filters and kernel size of 13, followed by 1 MaxPooling layer with a pool size of 2, followed by
1 Convolutional layer with 32 filters and kernel size of 3, followed by 1 MaxPooling layer with a pool size of 2, followed by
1 Flatten layers, followed by 2 Dense layers with 64 and 32 neurons, respectively, with 0.7 of Dropout, followed by 1 last dense layer with 1 neuron with a sigmoid function. Optimizer: Adam. Loss Function: Binary Crossentropy.

\noindent
Query3:
1 Convolutional layer with 32 filters and kernel size of 13, followed by 1 MaxPooling layer with a pool size of 2, followed by
1 Flatten layer, followed by 1 Dense layer with 64 neurons with 0.7 of Dropout, followed by 1 last dense layer with 10 neurons with a sigmoid function. Optimizer: Adam. Loss Function: Binary Crossentropy.

\noindent
Query4:
1 Convolutional layer with 64 filters and kernel size of 13, followed by 1 MaxPooling layer with a pool size of 2, followed by
1 Convolutional layer with 32 filters and kernel size of 3 followed by 1 MaxPooling layer with a pool size of 2, followed by
1 Flatten layer, followed by 1 Dense layer with 128 neurons with 0.5 of Dropout followed by 1 last dense layer with 10 neurons with a sigmoid function. Optimizer: Adam. Loss Function: Binary Crossentropy.

\noindent
Query5:
1 Convolutional layer with 64 filters and kernel size of 13, followed by 1 MaxPooling layer with a pool size of 2, followed by
1 Flatten layers followed by 1 Dense layer with 128 and 64 neurons, respectively, with 0.5 of Dropout, followed by 1 last dense layer with 1 neuron with a sigmoid function. Optimizer: Adam. Loss Function: Binary Crossentropy.

\noindent
Query6:
1 Convolutional layer with 64 filters and kernel size of 13, followed by 1 MaxPooling layer with a pool size of 2, followed by 1 Flatten layer, followed by 1 Dense layer with 64 neurons with 0.5 of Dropou,t followed by 1 last dense layer with 15 neurons with a sigmoid function. Optimizer: Adam. Loss Function: Binary Crossentropy.

\noindent
Query7:
1 Convolutional layer with 64 filters and kernel size of 13, followed by 1 MaxPooling layer with a pool size of 2, followed by
1 Flatten layer, followed by 2 Dense layers with 128 and 64 neurons, respectively, with 0.5 of Dropout, followed by 1 last dense layer with 10 neurons with a sigmoid function. Optimizer: Adam. Loss Function: Binary Crossentropy.

\noindent
Query8:
1 Convolutional layer with 64 filters and kernel size of 13, followed by 1 MaxPooling layer with a pool size of 2, followed by
1 Flatten layer, followed by 1 Dense layer with 64 neurons with 0.5 of Dropout, followed by 1 last dense layer with 15 neurons with a sigmoid function. Optimizer: Adam. Loss Function: Binary Crossentropy.


\subsection{Training Phase Results}

Each graph shows loss and accuracy for training and validation for the indicated number of epochs.

\begin{figure}
\begin{center}
\includegraphics[width=0.8\textwidth]{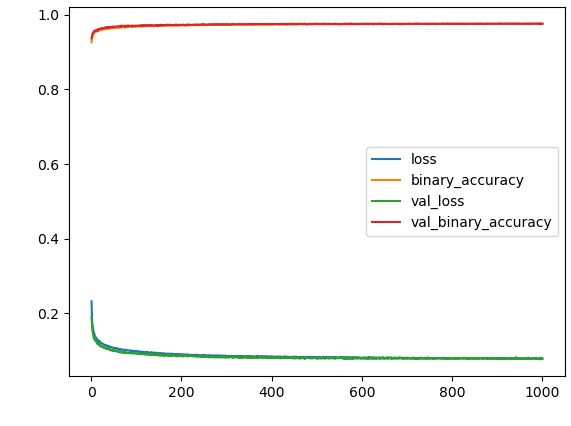}
\caption{Result of the LSTM training phase for the first Query}
\end{center}
\end{figure}

\begin{figure}
\begin{center}
\includegraphics[width=0.8\textwidth]{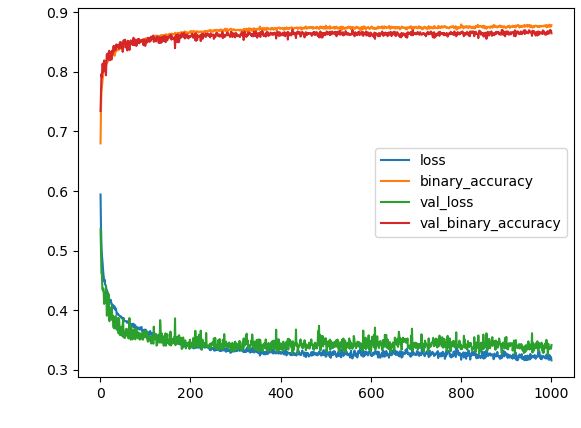}
\caption{Result of the LSTM training phase for the second Query}
\end{center}
\end{figure}

\begin{figure}
\begin{center}
\includegraphics[width=0.8\textwidth]{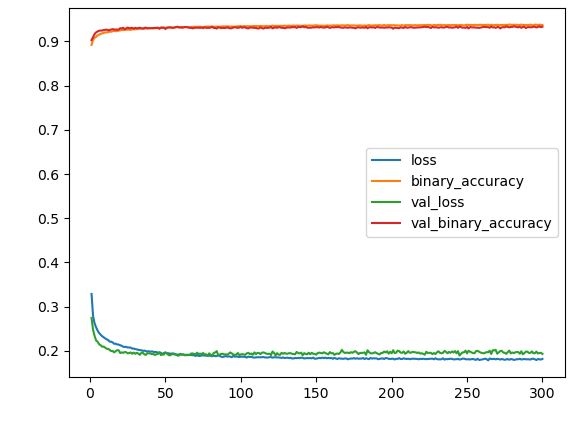}
\caption{Result of the LSTM training phase for the third Query}
\end{center}
\end{figure}

\begin{figure}
\begin{center}
\includegraphics[width=.8\textwidth]{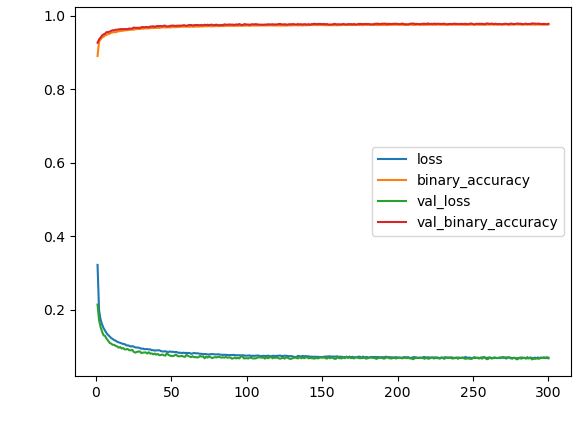}
\caption{Result of the LSTM training phase for the fourth Query}
\end{center}
\end{figure}

\begin{figure}
\begin{center}
\includegraphics[width=0.8\textwidth]{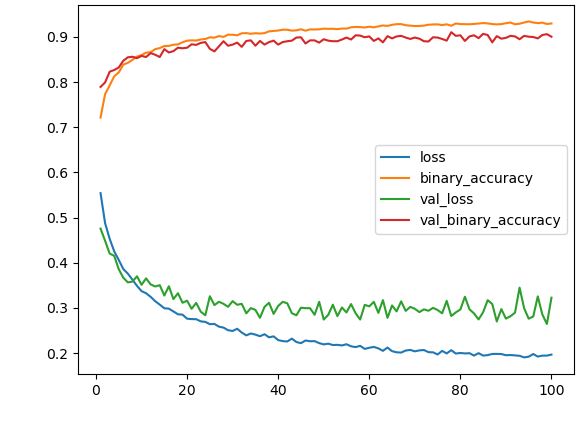}
\caption{Result of the LSTM training phase for the fifth Query}
\end{center}
\end{figure}

\begin{figure}
\begin{center}
\includegraphics[width=0.8\textwidth]{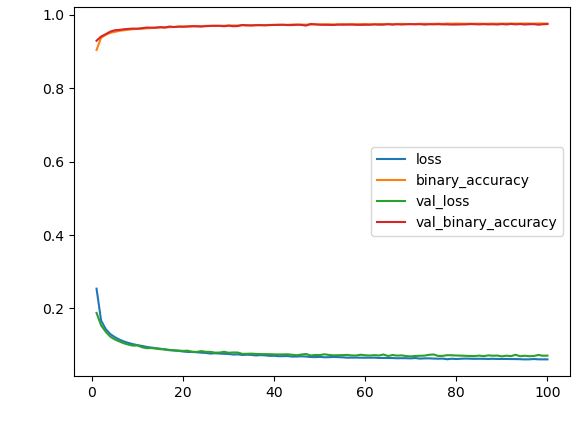}
\caption{Result of the LSTM training phase for the sixth Query}
\end{center}
\end{figure}

\begin{figure}
\begin{center}
\includegraphics[width=0.8\textwidth]{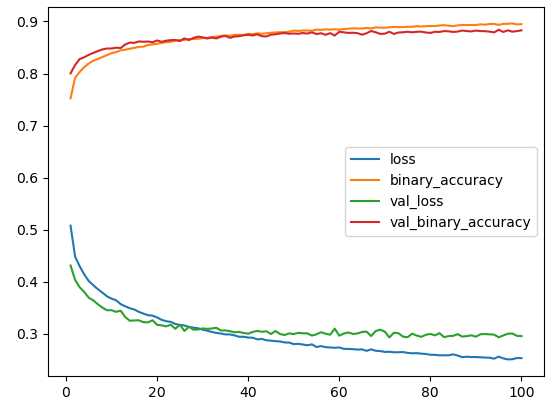}
\caption{Result of the LSTM training phase for the seventh Query}
\end{center}
\end{figure}

\begin{figure}
\begin{center}
\includegraphics[width=0.8\textwidth]{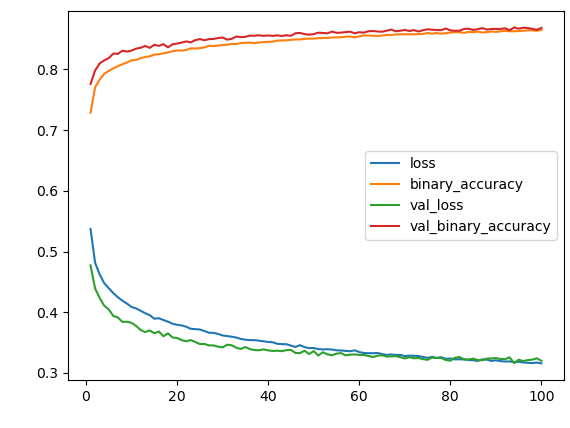}
\caption{Result of the LSTM training phase for the eighth Query}
\end{center}
\end{figure}


\begin{figure}
\begin{center}
\includegraphics[width=\textwidth]{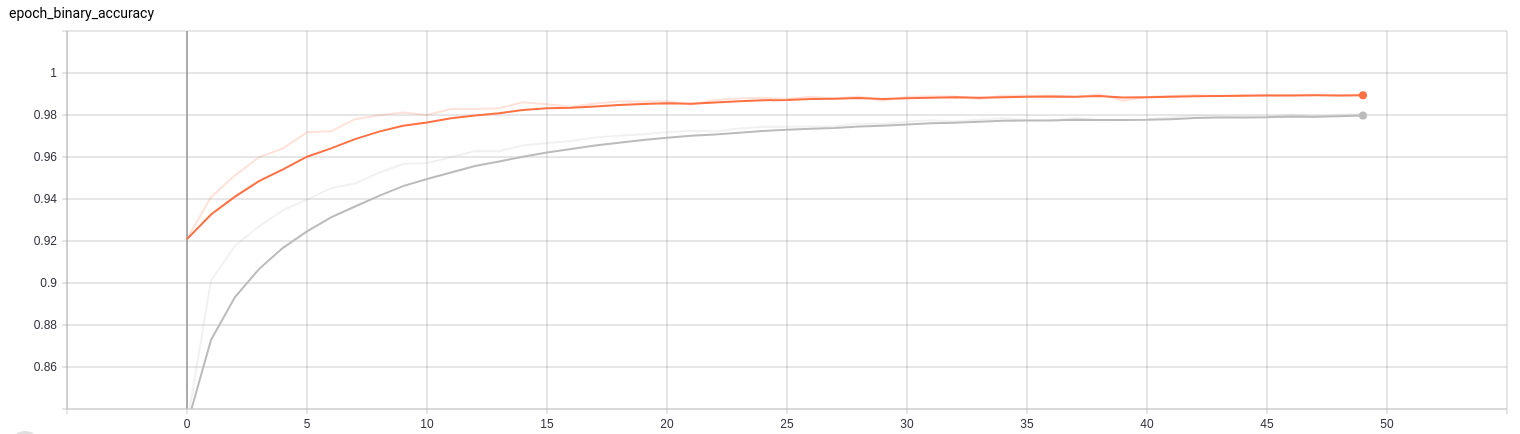}
\includegraphics[width=\textwidth]{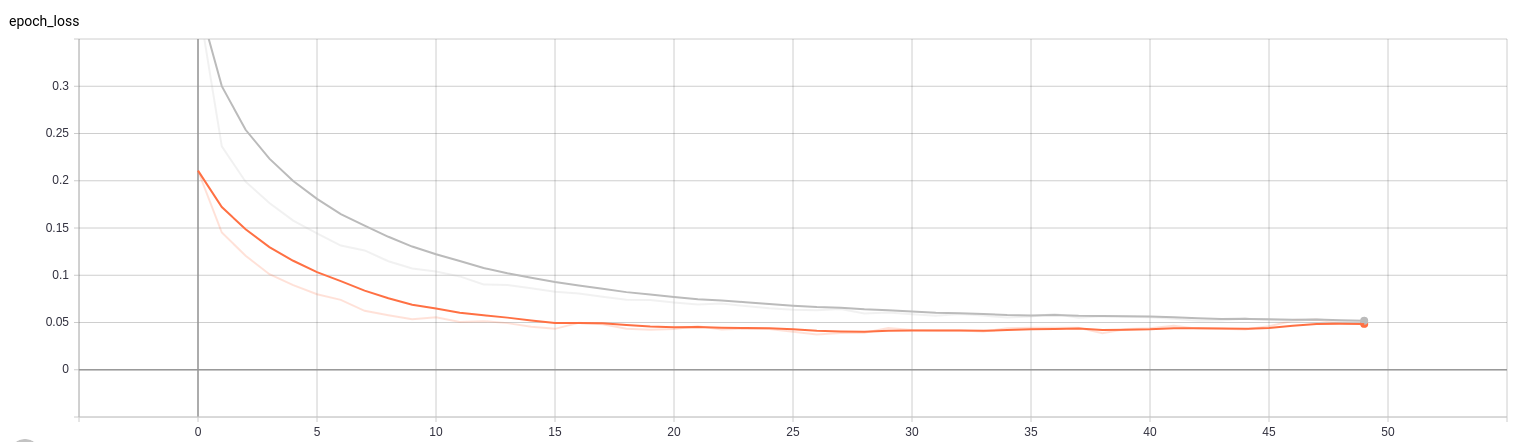}
\caption{Result of the CNN training phase for the first Query, with gray representing the training and orange the validation. }
\end{center}
\end{figure}

\begin{figure}
\begin{center}
\includegraphics[width=\textwidth]{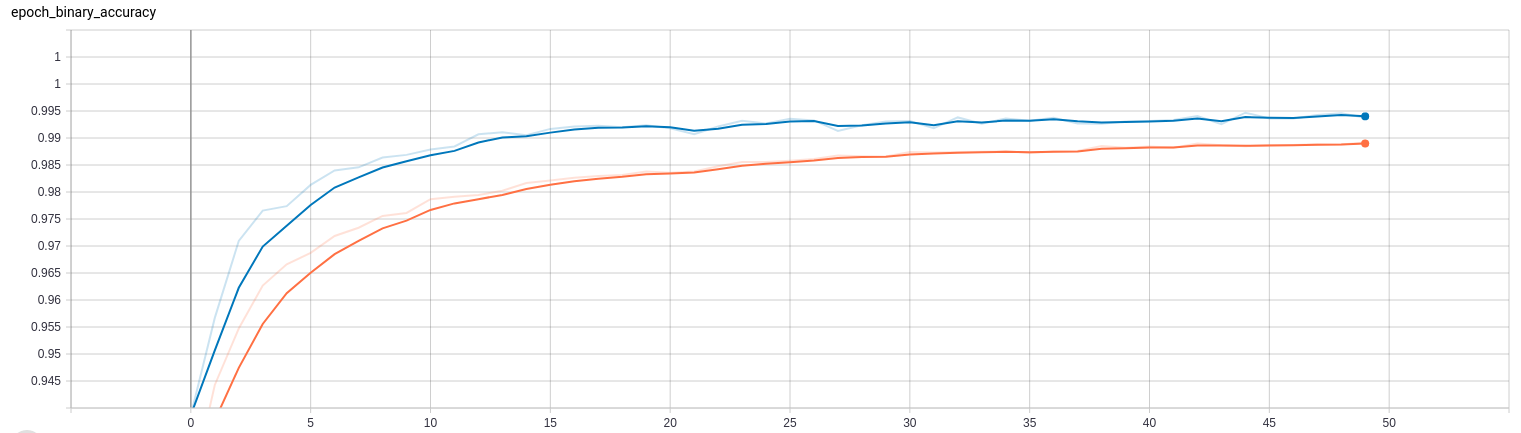}
\includegraphics[width=\textwidth]{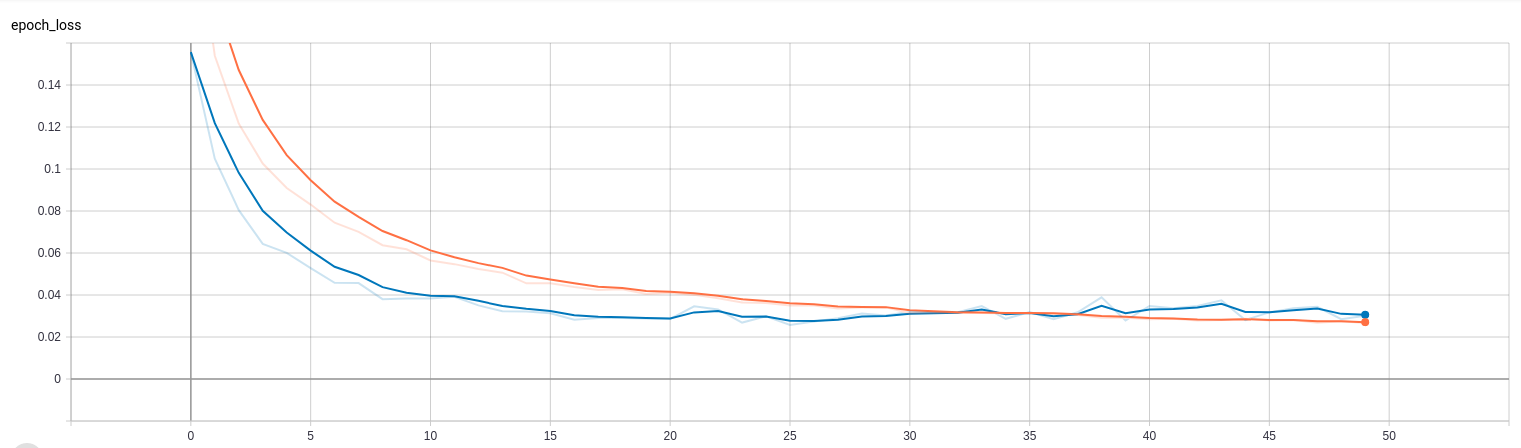}
\caption{Result of the CNN training phase for the second Query, with orange representing the training and blue the validation. }
\end{center}
\end{figure}

\begin{figure}
\begin{center}
\includegraphics[width=\textwidth]{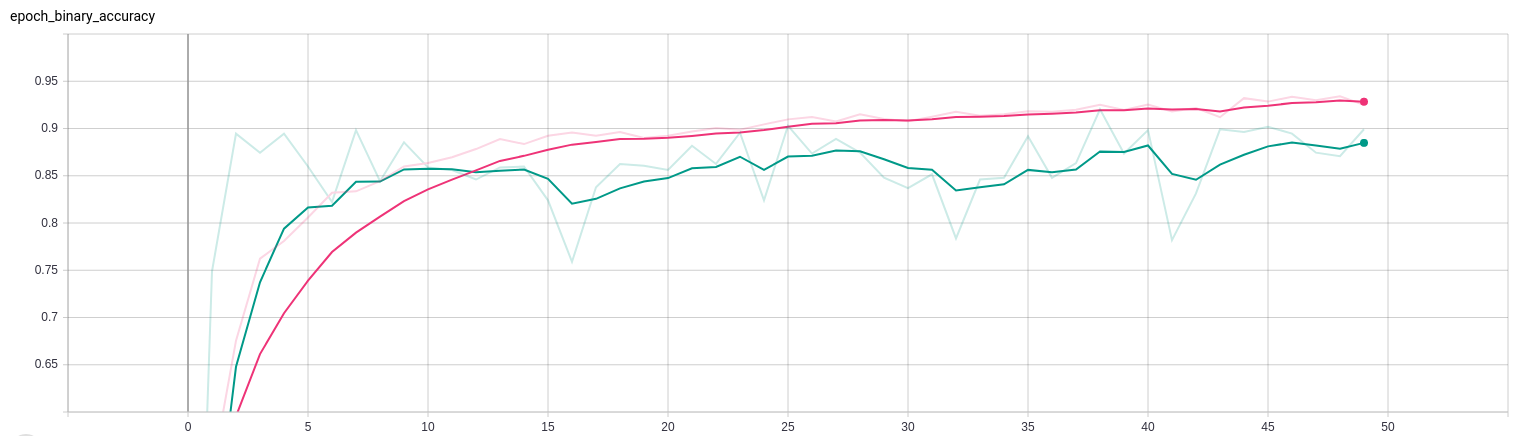}
\includegraphics[width=\textwidth]{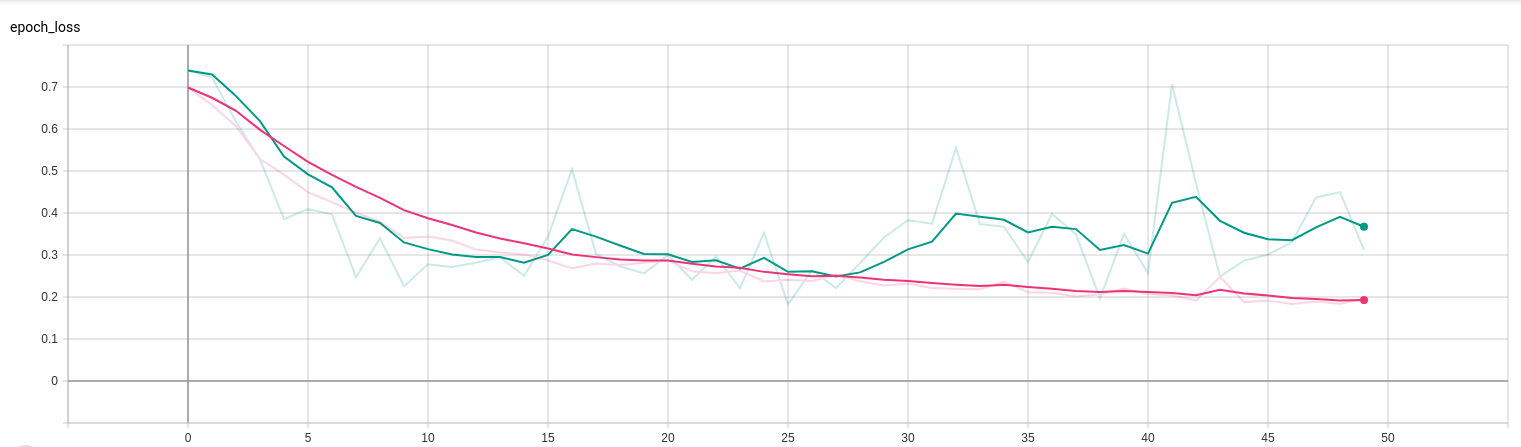}
\caption{Result of the CNN training phase for the third Query, with pink representing the training and green the validation. }
\end{center}
\end{figure}

\begin{figure}
\begin{center}
\includegraphics[width=\textwidth]{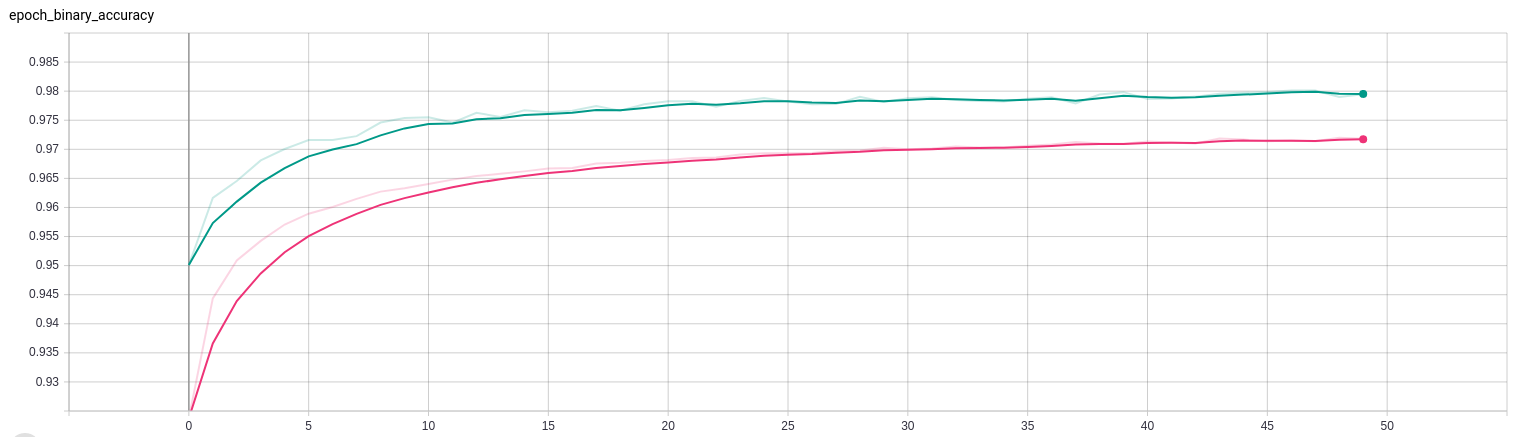}
\includegraphics[width=\textwidth]{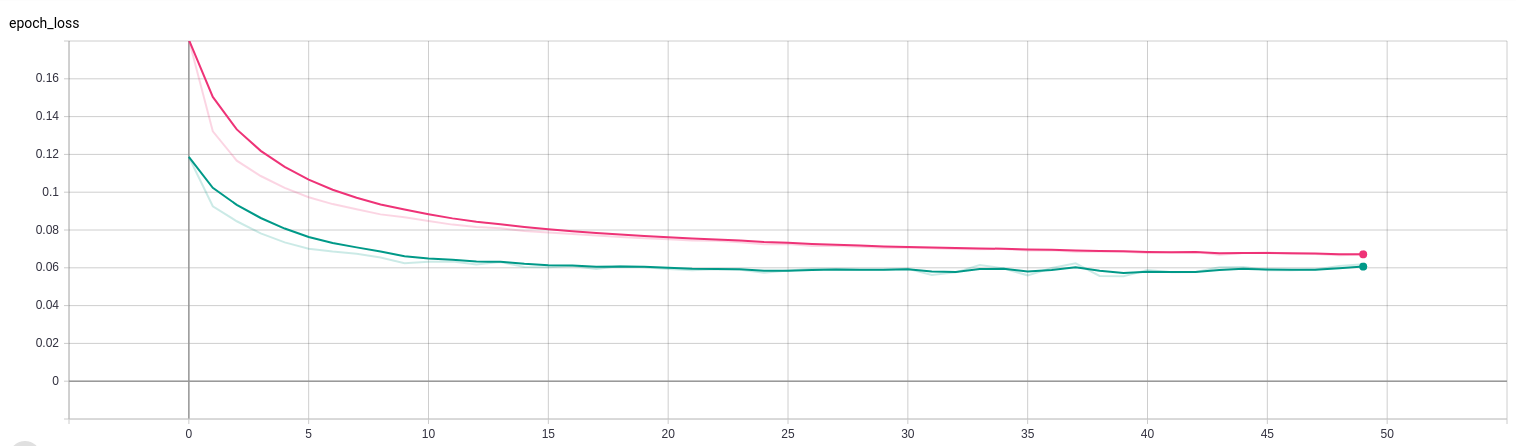}
\caption{Result of the CNN training phase for the fourth Query, with green representing the training and pink the validation. }
\end{center}
\end{figure}

\begin{figure}
\begin{center}
\includegraphics[width=\textwidth]{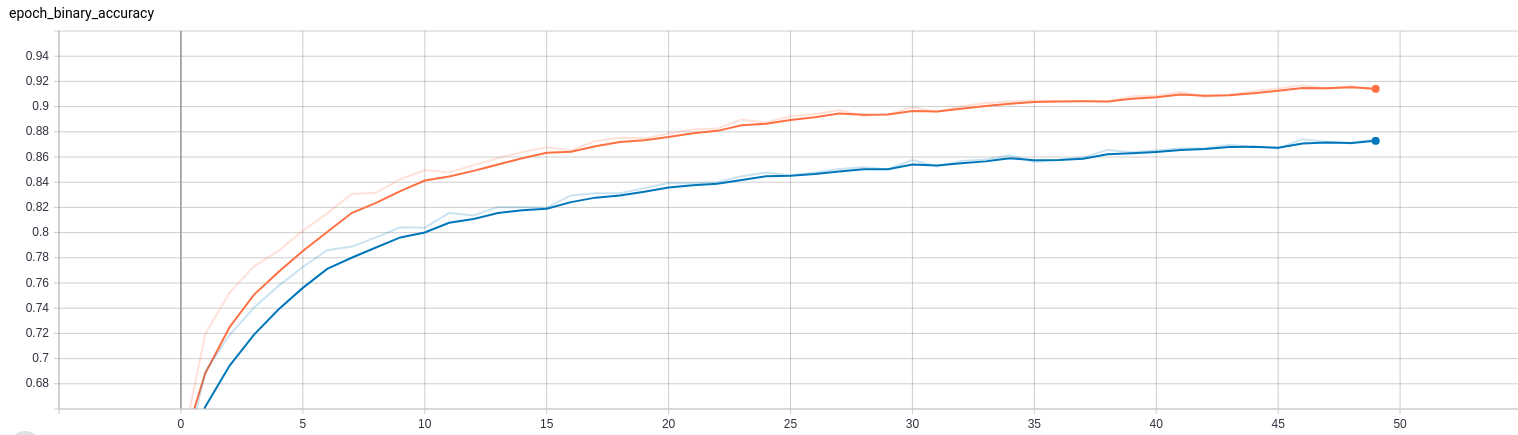}
\includegraphics[width=\textwidth]{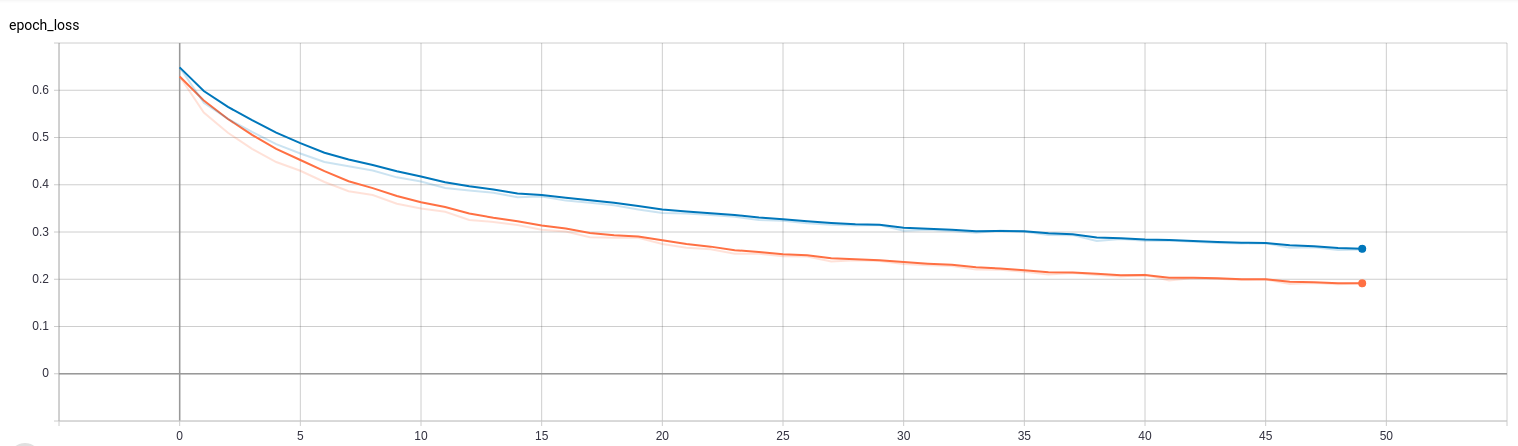}
\caption{Result of the CNN training phase for the fifth Query, with orange representing the training and blue the validation. }
\end{center}
\end{figure}

\begin{figure}
\begin{center}
\includegraphics[width=\textwidth]{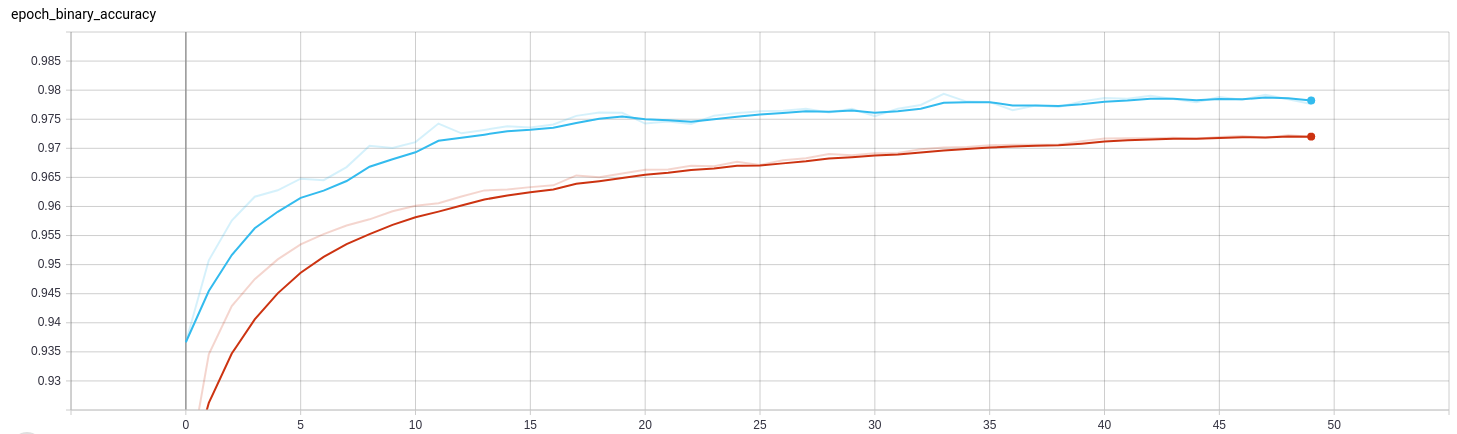}
\includegraphics[width=\textwidth]{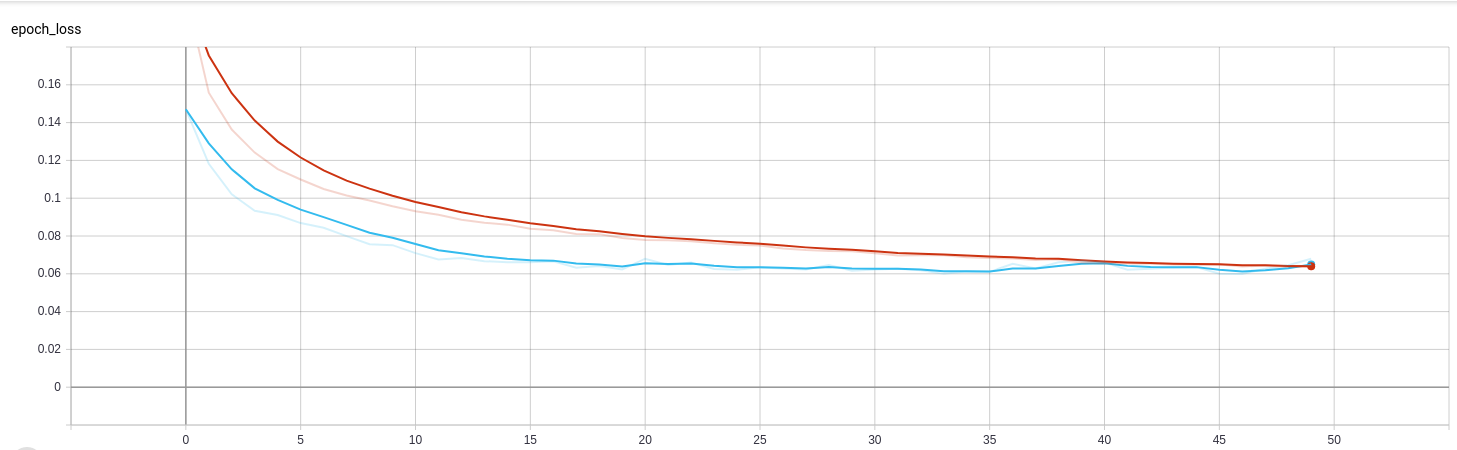}
\caption{Result of the CNN training phase for the sixth Query, with blue representing the training and red the validation. }
\end{center}
\end{figure}

\begin{figure}
\begin{center}
\includegraphics[width=\textwidth]{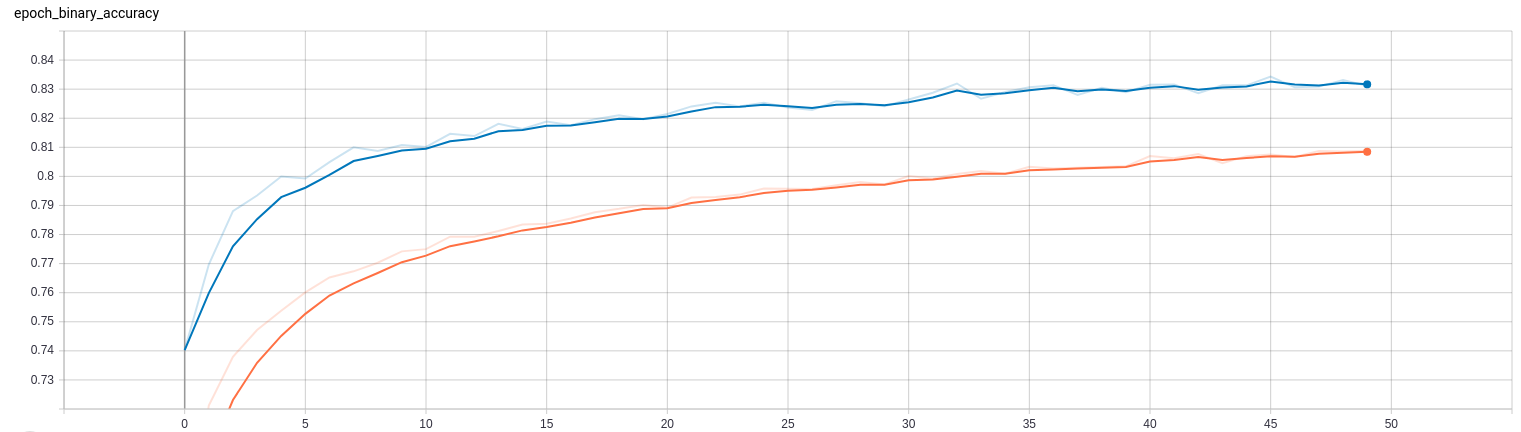}
\includegraphics[width=\textwidth]{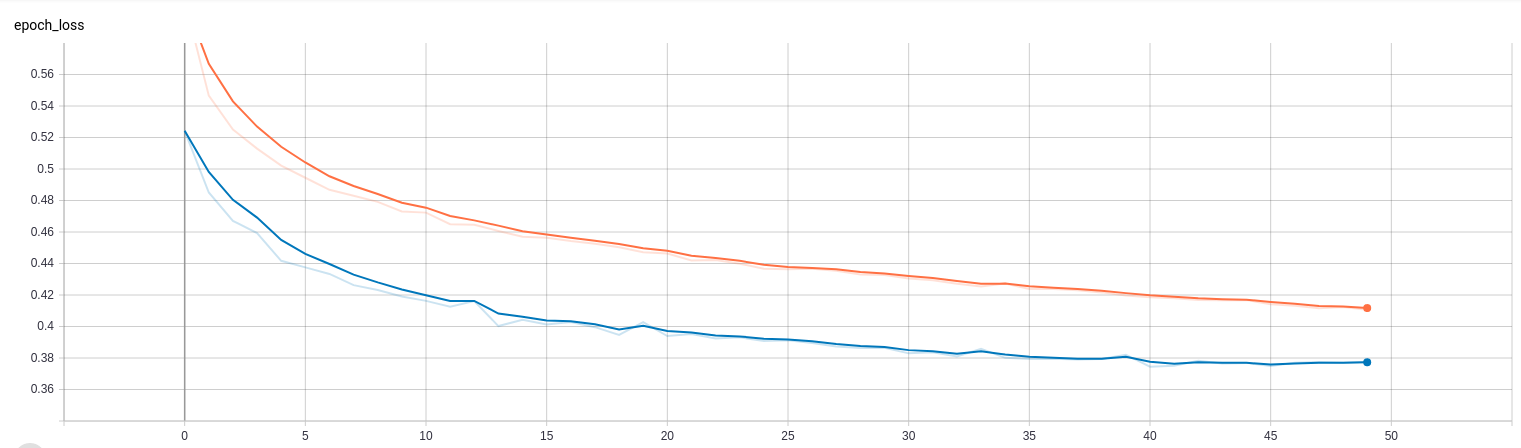}
\caption{Result of the CNN training phase for the seventh Query, with blue representing the training and orange the validation. }
\end{center}
\end{figure}

\begin{figure}
\begin{center}
\includegraphics[width=\textwidth]{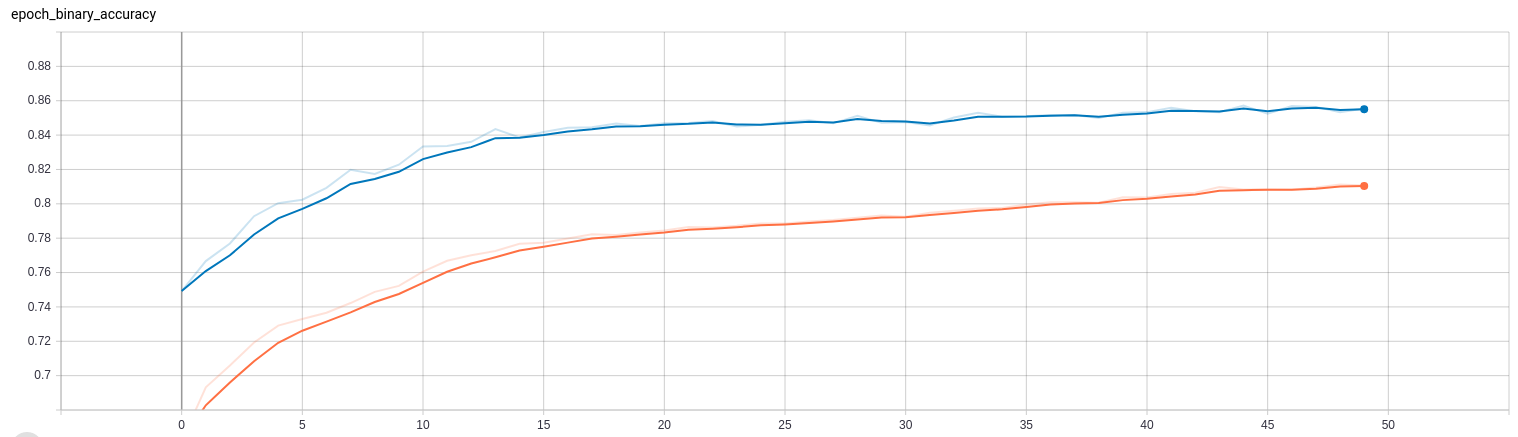}
\includegraphics[width=\textwidth]{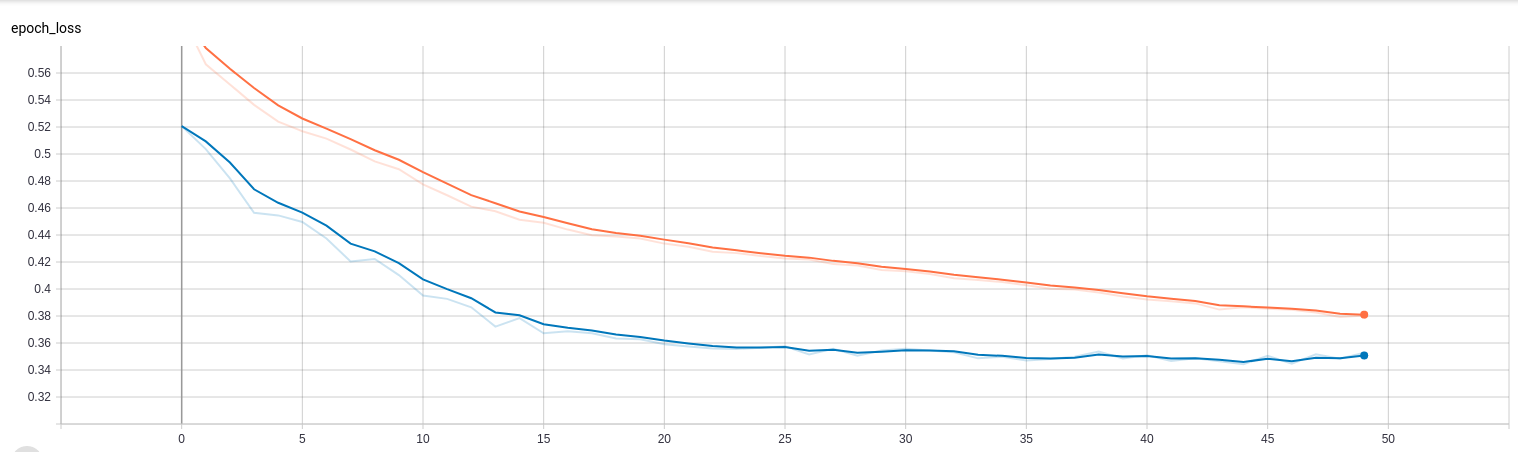}
\caption{Result of the CNN training phase for the eighth Query, with blue representing the training and orange the validation. }
\end{center}
\end{figure}

\end{document}